\documentclass{article}

\usepackage[final]{corl_2026} 

\usepackage{graphicx}
\usepackage{wrapfig}
\usepackage{xcolor}
\usepackage{colortbl}
\usepackage[font=small]{caption}
\usepackage{lipsum}
\usepackage{microtype}
\usepackage{makecell}
\usepackage{amsmath}
\usepackage{siunitx}
\usepackage{amssymb}
\usepackage{booktabs}
\usepackage{multirow}
\usepackage{enumitem}
\usepackage{arydshln}

\title{
FingerViP: Learning Real-World Dexterous Manipulation with Fingertip Visual Perception
}

\author{
\textbf{
Zhen Zhang\textsuperscript{1},
Weinan Wang\textsuperscript{1,*},
Hejia Sun\textsuperscript{2,*},
Qingpeng Ding\textsuperscript{1},
Xiangyu Chu\textsuperscript{1,4},
}\\
\textbf{
Guoxin Fang\textsuperscript{1,3},
K. W. Samuel Au\textsuperscript{1,4}
}
\\[0.3em]
{\normalfont\mdseries
\begin{tabular}{c}
\fontsize{9}{11}\selectfont
\textsuperscript{1}The Chinese University of Hong Kong
\quad
\textsuperscript{2}The Hong Kong Polytechnic University
\\[-0.1em]
\fontsize{9}{11}\selectfont
\textsuperscript{3}Institute of Intelligent Design and Manufacturing, CUHK
\quad
\textsuperscript{4}Multi-Scale Medical Robotics Center
\\
\fontsize{9}{11}\selectfont
\textsuperscript{*}Equal contribution
\end{tabular}
}
}
\begin{document}
\maketitle



\vspace{-4mm}
\begin{abstract}
Current dexterous manipulation systems commonly rely on external or wrist-mounted cameras for visual feedback. However, such sensing setups are prone to occlusion and can limit performance on tasks requiring close-range, multi-view perception. In this work, we design a vision-enhanced fingertip module that brings clear visual perception directly to the fingertips and present FingerViP, a learning system that leverages fingertip visual perception for dexterous manipulation. Specifically, the fingertip module integrates miniature-camera sensing within compact-footprint hardware and is installed on each finger of a multi-fingered hand, which substantially improves visual perception by providing comprehensive, multi-view observations of objects and surrounding environment. Building on the integrated fingertip modules, we develop a diffusion-based whole-body visuomotor policy conditioned on a third-view camera and multi-view fingertip vision, enabling the robot to effectively learn complex manipulation skills directly from human demonstrations. To improve view-proprioception alignment and contact awareness in the policy observation representation, each fingertip visual feature is augmented with its corresponding camera-pose encoding and per-finger joint-current encoding. We validate the effectiveness of multi-view fingertip vision and demonstrate the robustness and adaptability of FingerViP across a range of challenging real-world tasks, including retrieving objects behind an occluding curtain, achieving a $56.6\%$ relative improvement in average success rate over the baseline using an external third-view camera and a wrist-mounted camera. Videos, code, and hardware designs are available on our project page:~\url{https://fingervip.github.io/}.
\end{abstract}


\section{Introduction}
Dexterous manipulation~\cite{lin2024learning, zitkovich2023rt, sliwowski2025demonstrating, wu2025tacdiffusion, lin2024generalize} remains one of the most challenging yet fundamental problems in robotics. Recent advances in visuomotor policy learning for dexterous manipulation~\cite{wang2024dexcap, yang2025ace, cheng2025open} have highlighted the benefits of vision-based perception for improving environmental awareness and action generation. Many existing systems use either external cameras~\cite{fu2025cordvip, chen2025vividex, jain2019learning} or a wrist-mounted camera~\cite{wan2025rapid, hu20253d, chi2024universal}, while others~\cite{zhang2025kinedex, ni2025vo, chen2023sequential} combine both modalities. Despite their utility, views from external and wrist-mounted cameras can be unreliable in narrow, cluttered, and heavily occluded spaces. Specifically, external cameras are distant and frequently occluded by the hand and objects, obscuring contact-level interactions~\cite{jangir2022look,feng2025seeing}. In contrast, wrist cameras are closer yet limited by a single viewpoint and are often occluded by the hand and the manipulated object during interaction (e.g., grasping), hindering localization and contact monitoring~\cite{xu2025dexumi,nonnengiesser2025hand,zhu2025touch}. Moreover, the physical bulk of a wrist camera reduces clearance and restricts hand motion, which is particularly problematic in tight, confined settings. These limitations intuitively motivate a perception mechanism that provides close-range, multi-view observations without adding additional wrist-mounted hardware, while enabling dexterous manipulation particularly in confined environments.

To address this challenge, we design a vision-enhanced fingertip module that embeds a miniature camera in each fingertip of a dexterous hand. Compared with external and wrist-mounted cameras, fingertip vision provides clear, semantically rich RGB observations, and more importantly, a close-range view near contact, capturing fine-grained interaction cues (e.g., local texture) that are often occluded from distal viewpoints. This contact-proximal view also supports useful cues for inferring contact states (e.g., contact onset) during interaction, whereas vision-based tactile sensors typically require costly specialized hardware and remain limited to the contact patch~\cite{lambeta2020digit,yuan2017gelsight}. Equipping multiple fingertip modules further enables multi-view perception, yielding more comprehensive observations of both the hand and the surrounding scene. Moreover, the fingertip module preserves the hand’s form factor and avoids interference in tight-clearance manipulation, while providing rich semantic cues in confined and heavily occluded spaces. 

Building on our fingertip design, we propose a learning-based framework to train a visuomotor policy directly from human demonstrations, enabling a systematic evaluation of the advantages of fingertip visual perception for dexterous manipulation. To address the challenges posed by dynamic, high-dimensional multi-fingertip inputs with continuously changing viewpoints, we use a transformer-based diffusion model~\cite{chi2025diffusion} that conditions action denoising on fingertip visual tokens and robot proprioception. Additionally, to preserve the spatial information and contact-related signals of fingertip vision, fingertip-view poses and per-finger joint currents are encoded and concatenated to visual features. By fusing fingertip (hand-centric) observations with third-view (global) context, the policy leverages complementary cues for both contact-rich interaction and scene-level awareness, facilitating robust policy learning from human demonstrations. 

In summary, our main contributions are as follows: \textbf{(i)} We develop a vision-enhanced fingertip module with an embedded miniature camera and deploy it on each finger of a multi-fingered hand, enabling hand-centric multi-view observations that effectively mitigate occlusions without significantly increasing the hand’s physical envelope. \textbf{(ii)} We propose FingerViP, a learning system that leverages third-view and multi-view fingertip observations with fingertip-camera pose and per-finger joint-current encodings, to improve view–proprioception alignment and contact awareness, enabling robust dexterous manipulation. \textbf{(iii)} We demonstrate the effectiveness and generalization of FingerViP through a range of challenging real-world tasks and scenarios.

\begin{figure}[tb]
    \centering
    \includegraphics[width=1.0\textwidth]{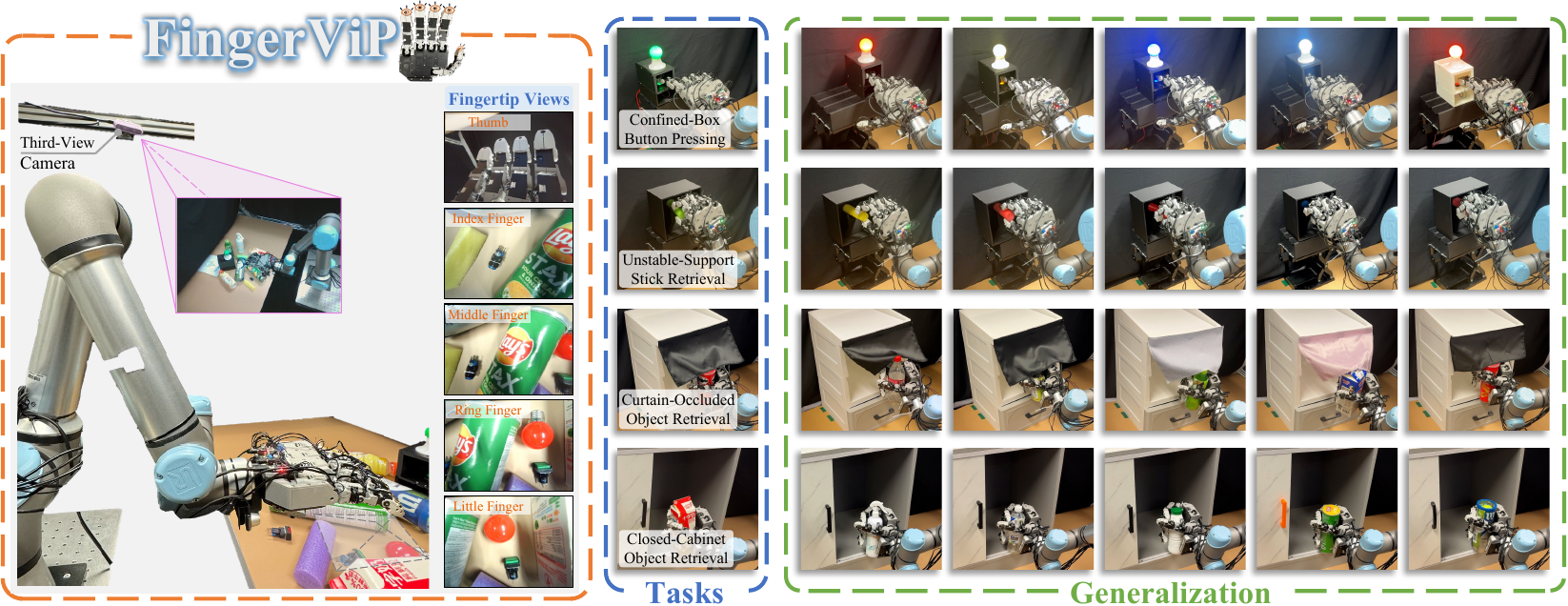}
    \caption{We present \textbf{FingerViP}, a learning system that leverages fingertip visual perception to improve dexterous manipulation, especially under confined and highly-occluded settings. FingerViP achieves robust performance across diverse real-world tasks and generalization scenarios.}
    \label{whole_frame}
    \vspace{-4mm}
\end{figure}

\section{Related Works}
\noindent\textbf{Perception Modalities for Dexterous Manipulation}\quad
Multi-fingered robotic systems rely primarily on tactile and visual feedback for complex dexterous manipulation. Recently, tactile sensors that mimic human skin have attracted increasing attention, enabling contact-rich manipulation~\cite{funabashi2022multi, wu2025tacdiffusion, huang2022soft, lin2025pp}. However, commonly used tactile sensors for robotic hands~\cite{yuan2017gelsight, lambeta2020digit, lin20239dtact, XELA_uSkin, BioTac} are often costly, structurally complex, or difficult to manufacture. More importantly, most provide feedback only after physical contact with an object~\cite{matak2022planning, zhao2025touch}, limiting their ability to perceive object position or scene context during pre-contact phases, such as approach and alignment.
Recent advances in visuomotor policy using visual observations~\cite{chi2025diffusion, zhao2023learning} have shown promising performance in end-to-end manipulation. Existing multi-fingered hand systems commonly use wrist-mounted cameras~\cite{wan2025rapid, fang2025dexop} and third-view cameras~\cite{andrychowicz2020learning, weng2024dexdiffuser, fu2025cordvip}, enabling robots to perceive pre-contact information and task-relevant scene context during long-horizon dexterous manipulation.

\noindent\textbf{Visuotactile and Proximity Sensing}\quad
Optical tactile sensors~\cite{yuan2017gelsight, lambeta2020digit, lin20239dtact} capture coated elastomer deformation for high-resolution contact sensing, but are not designed to provide clear  RGB images. Recent visuotactile and multimodal fingertip sensors integrate tactile and proximity sensing for robotic manipulation. FingerVision~\cite{fingervision, Imptactile} and Finger-STS~\cite{Finger-STS, Multimodal} use see-through soft skins and visual markers to enable optical tactile sensing and proximity perception, but deformable skins and marker patterns can compromise visual clarity or reduce the availability of clean semantic RGB observations. Other designs, such as selectively transmissive membranes~\cite{Membrane}, mechanical switching sensors~\cite{dong2026look}, and polarization-based sensors~\cite{Vi2TaP}, improve multimodal perception but introduce additional optical or mechanical complexity, and may incur switching latency. StereoTac~\cite{StereoTac} and VisTac~\cite{VisTac} further enhance local geometry or tactile perception through semi-transparent coatings, controllable illumination, or multi-camera structures,  while HydroPalm~\cite{HydroPalm} and the multimodal fingertip sensor in~\cite{saloutos2022design} focus more on tactile/force/proximity perception rather than high-quality RGB fingertip vision. Overall, it remains challenging for existing multimodal sensors to tightly integrate into multi-fingered hands while simultaneously providing high-quality fingertip RGB observations, due to their optical structures, fabrication complexity, bulky form factors, and limited visual clarity.

\noindent\textbf{Imitation Learning from Demonstrations}\quad
Imitation learning (IL) provides a direct way for robots to acquire manipulation skills from expert demonstrations. Early behavioral cloning (BC) formulations~\cite{bojarski2016end, lin2021limitations, guss2021towards, calinon2007learning} are often insufficient to capture fine-grained multimodal variations in human demonstrations~\cite{pearce2023imitating}. Other IL paradigms~\cite{ziebart2008maximum, ho2016generative, ross2011reduction} typically involve high interaction costs, computational complexity, and limited scalability in real-world robotic systems~\cite{hussein2017imitation, osa2018algorithmic}. Over the past few years, Denoising Diffusion Probabilistic Models (DDPMs)~\cite{ho2020denoising} have become a powerful framework for generative modeling and robotic control. Recent works have extended diffusion models to motion planning~\cite{carvalho2023motion}, trajectory optimization~\cite{pan2024model}, offline reinforcement learning~\cite{janner2022diffuser, pearce2023imitating}, and visuomotor policy learning~\cite{chi2025diffusion}. More recent advances~\cite{zhu2025scaling, ma2024hierarchical, ze20243d, hou2025dita, xu2025robopanoptes} further explore the potential of diffusion-based robot behavior generation and dexterous manipulation with multi-fingered hands~\cite{weng2024dexdiffuser, fu2025cordvip}, demonstrating promising performance in real-world tasks. Overall, these studies suggest that diffusion models provide a scalable and expressive framework for robust robot behavior generation.

\section{Methods}
\begin{wrapfigure}{R}{0.5\linewidth}
    \centering
    \vspace{-2mm}
    \includegraphics[width=1.0\linewidth]{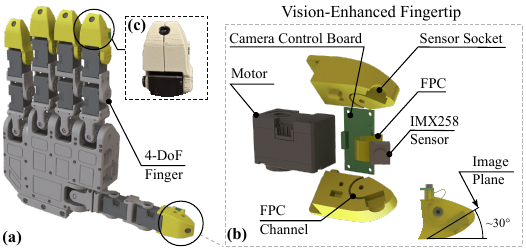}
    \caption{\textbf{Vision-enhanced fingertip module.} (a) The 20-DoF dexterous hand. (b) Exploded CAD view of the fingertip module. (c) Fabricated and assembled fingertip module.}
    \label{dexterous_hand}
    \vspace{-2mm}
\end{wrapfigure}

\textbf{Robot Dexterous Hand}\quad
As shown in Fig.~\ref{dexterous_hand} (a), the robotic hand setup used in this work is inspired by the open-source RAPID Hand~\cite{wan2025rapid}. The hand is 3D-printed and servo-actuated, with an anthropomorphic five-finger architecture comprising 20 DoFs, four for each finger. Building on this platform, we developed a vision-enhanced fingertip module.

\subsection{Vision-Enhanced Fingertip Module}
The vision-enhanced fingertip module embeds a miniature camera in a low-cost, compact, and highly modular design, enabling quick fabrication, easy integration, and high maintainability.

\noindent\textbf{Fingertip Camera Selection}\quad
Three key criteria were considered to select the fingertip cameras: (i) physical size, (ii) field of view (FOV), and (iii) connectivity. Given the limited fingertip volume, we chose a low-cost \texttt{Generic 12MP SONY IMX258 USB Camera Module}, which features a split sensor--PCB design connected by a short flexible printed circuit (FPC). Its compact sensor package ($8.5\times8.5\times4.95\;\si{\milli\meter}$) enables easy integration within the fingertip. The controller PCB dimensions ($30\times15\times1.6\;\si{\milli\meter}$) allow it to be mounted externally on the dorsal side while preserving a compact and modular fingertip design. The camera provides a wide $135^\circ$ FOV, and its USB Video Class-compliant nature allows for plug-and-play over USB with no additional drivers.

\noindent\textbf{Structural Modification and Camera Integration}\quad
The sensor is integrated into the fingertip within a pocket tilted by $\sim30^\circ$ relative to the palmar surface so that the camera can jointly observe the region ahead of and beneath the finger, as shown in Fig.~\ref{dexterous_hand} (b). The fragile FPC--sensor junction is embedded inside the fingertip body, and the PCB is mounted externally on the dorsal side (fingernail position), improving the reliability of the FPC-to-PCB connection while keeping the fingertip compact and modular. Concretely, an internal S-shaped channel is kept to route the FPC. During assembly, the sensor and FPC are fixed in one half-shell and fully encapsulated when the second half is closed, assisted by the mortise–tenon alignment. The fabricated and assembled module is shown in Fig.~\ref{dexterous_hand} (c).

\subsection{Teleoperation and Data Collection}
To efficiently collect human demonstrations, we employ a high-DoF teleoperation interface~\cite{park2024avp} that captures human hand poses with an Apple Vision Pro and enables real-time control of the robotic arm and hand. In this work, we collect (i) target joint commands, (ii) robot proprioception, including arm-hand joint positions for all 26 DoFs and hand joint currents for the 20 hand DoFs, and (iii) multi-view RGB observations from six cameras: five fingertip cameras and one third-view RGB-D camera (RealSense D435i). In addition, synchronized wrist-view observations are recorded for baseline comparisons. More details are provided in Appendix~\ref{app:tdc}.

\subsection{Dexterous Visuomotor Policy}
\label{sec:policy}

\begin{figure}[tb]
    \centering
    \includegraphics[width=1.0\textwidth]{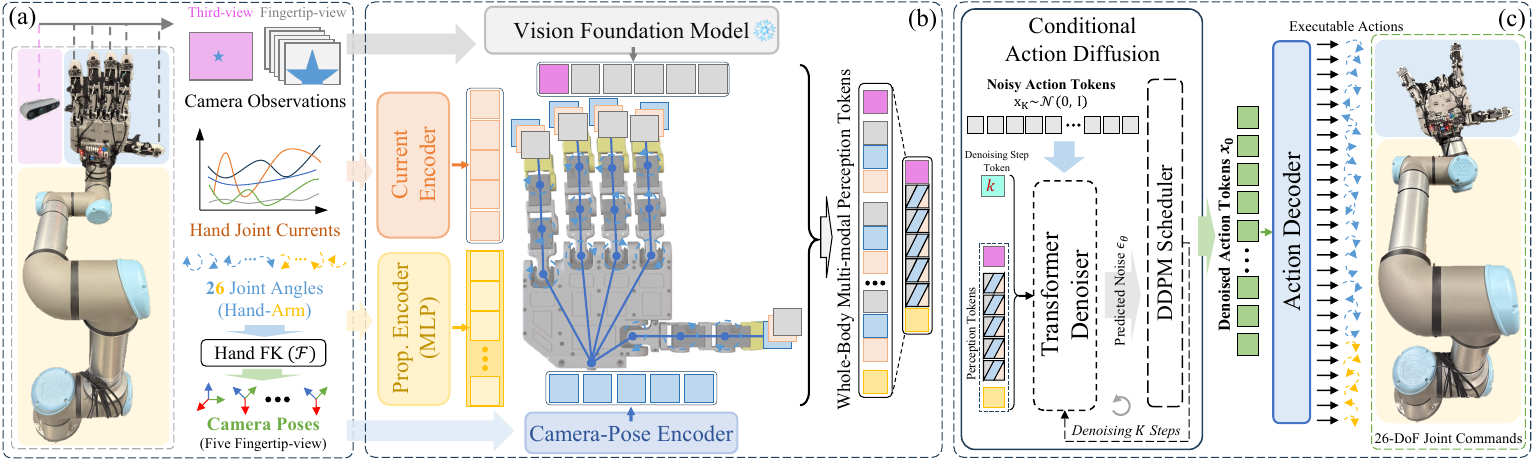}
    \caption{\textbf{Overview of FingerViP visuomotor policy with fingertip visual perception.} Multi-view visual features, fingertip-camera poses, finger joint currents, and robot proprioception are encoded as observation tokens to condition transformer-based action diffusion.
    }
    \label{policy}
    \vspace{-5mm}
\end{figure}

To control a 26-DoF robotic system, a transformer-based visuomotor policy is introduced and trained on human demonstrations to infer six joint angles of the arm and twenty joint angles of the hand simultaneously from six visual observations and proprioception. An overview (Fig.~\ref{policy}) of the policy is first introduced, followed by detailed explanations of the concrete modules.

\noindent\textbf{Policy Overview}\quad We formulate whole-body action prediction as a conditional denoising process following Diffusion Policy~\cite{chi2025diffusion}. At each time step $t$, the policy uses a short observation history, consisting of $O_{t-1}$ and $O_t$, to generate a future joint-command trajectory $A_t=\{a_t,a_{t+1},\ldots,a_{t+n-1}\}$. Following a receding-horizon control strategy, only the first $m$ actions are applied to the robot before the policy is queried again, where $m\leq n$. In our experiments, we set $m=8$ and $n=16$, following common practice. The observation $O_t$ contains both visual and proprioceptive information: $O_t=\{I_t,P_t,C_t,Q_t\}$. Here, $I_t$ denotes the RGB observations from one third-view camera and five fingertip-mounted cameras, with the resolution of $224\times224$. The pose set $P_t$ contains the hand-centric poses of the five fingertip cameras. The proprioceptive inputs include the whole-hand joint-current vector $C_t \in \mathbb{R}^{20}$ and the arm-hand joint-position vector $Q_t \in \mathbb{R}^{26}$. The action $a_t$ is represented as a 26-dimensional joint-space command covering the 6-DoF robot arm and the 20-DoF dexterous hand. During training, these action targets are obtained from teleoperation demonstrations.

\noindent\textbf{Coordinating Fingertip Views With Positional and Joint Current Encoding}\quad In FingerViP, the fingertip cameras are rigidly mounted on articulated fingers, so their image observations are naturally tied to the current hand configuration. The visual feature extracted from a fingertip image is therefore not only view-specific but also finger-state-dependent. To expose both the view-dependent image context and kinematics context to the policy, we attach a camera-pose embedding to each fingertip visual token, then aggregate it with finger joint currents. Inspired by the positional encoding strategy in~\cite{xu2025robopanoptes}, we compute the pose of each fingertip camera with respect to the hand base using forward kinematics and linearly project the camera pose to an embedding for each fingertip view. Specifically, we use Pinocchio to parse the hand URDF, identify individual links of the robot hand, and compute each fingertip camera pose in the hand base frame via the forward kinematics model $\mathcal{F}$ from the current hand joint angles $Q^{h}_{t}\subset Q_t$. Let ${}^{B}\mathbf{T}_{c, t}\in SE(3)$ denote the transformation from the hand base frame $B$ to camera frame $c$ at time step $t$: ${}^{B}\mathbf{T}_{c, t}=[{}^{B}\mathbf{R}_{c, t}\; {}^{B}\mathbf{p}_{c, t};\; \mathbf{0}^\top\; 1] = \mathcal{F}(Q^{h}_{t}),$ where ${}^{B}\mathbf{p}_{c, t}\in\mathbb{R}^{3}$ is the camera position and ${}^{B}\mathbf{R}_{c, t}\in SO(3)$ the orientation. Following the rotation representation of~\citet{zhou2019continuity}, we parameterize orientation by the first two columns of the rotation matrix, ${}^{B}\mathbf{r}_{c, t}=\mathrm{vec}\!\left({}^B\mathbf{R}_{c, t}[:,:\!2]\right)\in\mathbb{R}^{6},$ so that each camera pose is parameterized by a $3$-D position vector and a $6$-D rotation vector and represented by $( {}^{B}\mathbf{p}_{c, t}, {}^{B}\mathbf{r}_{c, t})\in\mathbb{R}^9$. The camera position and orientation are independently projected to $128$-D features using linear layers and concatenated into a $256$-D pose embedding. Beyond vision information, finger joint currents provide complementary signals that are tightly coupled to manipulation phases. During contact-rich phases such as grasping, holding, or pressing, joint currents often correlate with changes in loading and contact conditions, providing complementary contact-related cues that are not directly observable from RGB images alone under occlusion. Therefore, per-finger joint currents are included as contact-aware features. Rather than encoding the whole-hand current vector, we project each finger’s $4$-D current vector independently to a $128$-D feature via a linear layer to preserve finger-wise independence.

\noindent\textbf{Transformer-based Diffusion Policy Learning}\quad
For action generation, we use a transformer-based diffusion policy~\cite{chi2025diffusion} to predict a future joint-space action trajectory. During each denoising iteration $k$, the transformer denoiser refines noisy action tokens $\mathbf{x}_t^k$ by querying the observation-token sequence through cross-attention~\cite{vaswani2017attention}. After $K$ denoising steps, the resulting denoised tokens $\mathbf{x}_t^0$ are decoded into the future joint-command trajectory $A_t$. The observation tokens are constructed by combining multi-view visual features and proprioceptive embeddings. Specifically, we use a frozen CLIP-pretrained ViT-B/16 encoder~\cite{radford2021learning,dosovitskiy2020image} to extract visual features from the five fingertip cameras and the third-view camera. Prior robotic manipulation works have shown that vision foundation models such as CLIP can provide effective representations for challenging vision-based manipulation tasks~\cite{chi2024universal,xu2025flow, xu2025robopanoptes}. The five fingertip-view features are projected by a shared linear layer, whereas the third-view feature is processed by a separate (view-specific) projection layer. For each fingertip camera, the projected visual feature is concatenated with the corresponding camera-pose embedding and finger-current embedding to form a fingertip observation token. The third-view feature is used as a global visual token. Finally, the embedded robot joint positions are appended to the visual tokens to construct the observation-token sequence at time step $t$ for action prediction.

\section{Experiments}
To evaluate FingerViP, we conduct a set of challenging real-world dexterous manipulation tasks. To ensure fair comparisons, all methods for each task are trained using the same set of demonstrations and evaluated under identical initial robot states and scene configurations.

\subsection{Experiment Setup}
\noindent\textbf{Hardware}\quad We equipped a UR5e arm with the 20-DoF hand and used an Intel RealSense D435i as the static third-view camera. For baseline comparisons, we additionally mounted a D435i on the wrist to record synchronized wrist-view observations during teleoperation.

\noindent\textbf{Comparisons}\quad The primary focus of this work is to underscore the ability of the proposed system with fingertip visual perception. To this end, we compare our policy against the following conditions across all tasks: (I) \textit{Wrist Camera (WC)}: Observations from a single wrist-mounted camera. (II) \textit{Third-View Camera (TVC)}: Observations from a single third-view camera. (III) \textit{Fingertip Cameras Only (FTC)}: Observations from five fingertip cameras. (IV) \textit{TVC w/ WC}: Observations from both third-view camera and wrist-mounted camera. (V) \textit{FTC w/ WC}: Observations from five fingertip cameras and a wrist-mounted camera. Ours (FingerViP) uses both the third-view camera and five fingertip cameras, together with fingertip-camera pose encoding and per-finger joint-current encoding. For a fair comparison among methods using fingertip vision, FTC and FTC w/ WC use the same per-finger joint-current encoding as Ours.

\subsection{Real-world Tasks}
\begin{figure}[tb]
    \centering
    \includegraphics[width=1.0\linewidth]{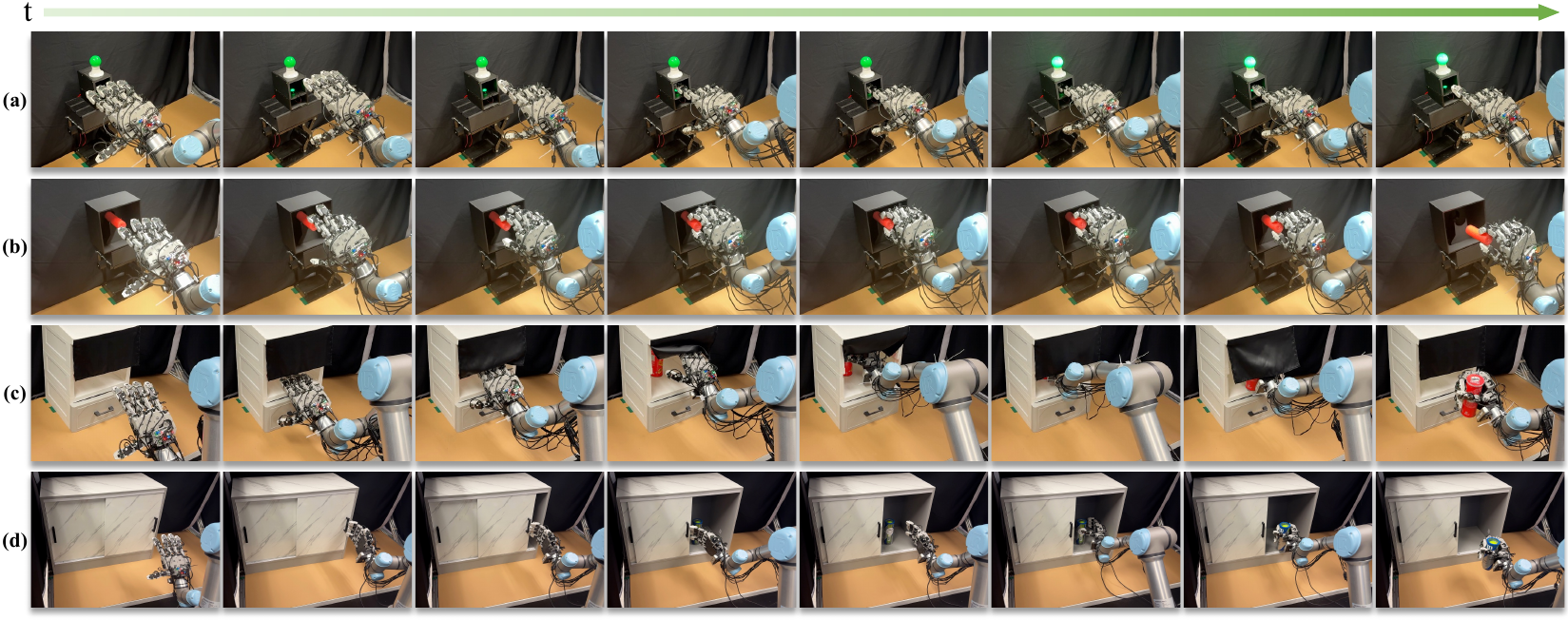}
    \caption{\textbf{Examples of the four dexterous manipulation tasks.} The pictures on the far left and right show the initial and final states, respectively. Each row shows the task progression over time.}
    \label{qualitative_results}
    \vspace{-4mm}
\end{figure}

We validate FingerViP in real‑world settings across four challenging dexterous manipulation tasks. Four representative task executions are shown in Fig.~\ref{qualitative_results}. Details of each task are provided in the following sections. Please refer to Appendix~\ref{app:scenarios} for more details of the testing scenarios.

\noindent\textbf{Confined-Box Button Pressing (S1)}\quad The task requires the robot to press a button inside a box, where the space allows only a single finger to act. The box with a colored illuminated button at the bottom is open on one side, and a bulb with the same color on the top serves as the success indicator, as shown in Fig.~\ref{qualitative_results} (a). Task success is indicated only by whether the bulb is lit. \textbf{Challenges.}\quad \underline{Highly-constrained space:} The confined space inside the box requires fine alignment and allows only one finger to reach the button and execute the press action. \underline{Visual occlusions:} The wrist-mounted and third-view cameras are nearly occluded when the other four fingers curl up and the extended finger enters the box. Once entered, the target button can only be observed and located by the index fingertip camera. \underline{High precision and contact sensitivity:} Task success requires (i) precise approach and alignment between the index finger and the small box opening, and (ii) contact-aware control to execute the press, avoiding slipping or missing the button. 



\noindent\textbf{Unstable-Support Stick Retrieval (S2)}\quad The task requires the hand to use three fingers, the thumb, index, and middle finger, to retrieve a lightweight foam craft stick, which is loosely supported by a slender U-shaped cradle, as shown in Fig.~\ref{qualitative_results} (b). Task success is defined as fully pulling the stick out from the support without dropping it. \textbf{Challenges.}\quad \underline{Constrained space:} The robot can only place three extended fingers inside the box to avoid collision with the wall. \underline{Unstable support:} The robot needs to contact and grasp the foam stick carefully without knocking it off the slender cradle. \underline{Multi-contact coordination:} Successful retrieval requires coordinated, stable contacts across multiple fingers during grasping and pulling; insufficient or unstable contact can cause the stick to slip or drop. \underline{Precision requirements:} The three fingers must maintain an open shape configured to center the stick, then accurately close the fingertips to conform to the stick surface for a secure grasp. 



\noindent\textbf{Curtain-Occluded Object Retrieval (S3)}\quad In this task, the robot is required to retrieve an object from a cabinet partially occluded by a hanging curtain, as shown in Fig.~\ref{qualitative_results} (c). Task success is defined as removing the object from the cabinet without it falling down. \textbf{Challenges.}\quad \underline{Local perception:} The object is placed inside the cabinet behind the hanging curtain, making local perception critical for guiding the hand to localize and grasp the target object. \underline{Visual occlusions:} Due to the hanging curtain, the object can only be observed through an opening at the bottom of the cabinet side. The top-down third-view camera can hardly observe the target object, and interaction with the curtain will block the other camera views during the robot execution phase. 

\noindent\textbf{Closed-Cabinet Object Retrieval (S4)}\quad In this long-horizon task, the robot is required to retrieve a target object from a closed cabinet, as shown in Fig.~\ref{qualitative_results} (d). In this task, we define two success metrics: successfully opening the cabinet and successfully retrieving the object without collision. \textbf{Challenges.}\quad \underline{Long-horizon manipulation:} First, the robot gently makes contact and slides the cabinet door, followed by pushing the door open to create sufficient clearance for safe entry. Second, the robot reaches into the cabinet, grasps the target object, and retrieves it. \underline{Local perceptual capability:} Given the handle’s small size, the robot needs fine-grained, close-range perception to accurately localize the handle and establish contact. \underline{Precision and contact awareness:} Door opening requires precise position control together with contact-aware control to maintain stable contact with the door. The robot must regulate the pushing distance and interaction force to generate enough clearance while avoiding damage to the cabinet. \underline{Visual occlusions:} The target object inside the cabinet is often heavily occluded and may be difficult to observe from a fixed top-down third-view camera. 
\begin{table}[tb]
    \centering
    \caption{Comparison with baselines across diverse real-world tasks.}
    \label{quantitative_results}
    \scriptsize
    \setlength{\tabcolsep}{7pt}  
    \begin{tabular}{lccccc}
        \toprule
        \multirow{2}{*}{\textbf{Methods}}
        & \multirow{2}{*}{\makecell{Confined-Box\\Button Pressing}}
        & \multirow{2}{*}{\makecell{Unstable-Support\\Stick Retrieval}}
        & \multirow{2}{*}{\makecell{Curtain-Occluded\\Object Retrieval}}
        & \multirow{2}{*}{\makecell{Closed-Cabinet \\ Object Retrieval}}
        & \multirow{2}{*}{\makecell{\textbf{Average}\\ \textbf{Success Rate}}} \\
        &  &  &  &  &  \\
        \toprule
        TVC        & $8/42$ $(19.0\%)$ & $14/37$ $(37.8\%)$ & $20/50$ $(40.0\%)$ & $ 30/55$ $(54.5\%)$ & $37.8\%$  \\
        WC         & $2/42$ $(4.8\%)$ & $10/37$ $(27.0\%)$ & $28/50$ $(56.0\%)$ & $ 34/55$ $(61.8\%)$ & $37.4\%$  \\
        TVC w/ WC  & $9/42$ $(21.4\%)$ & $17/37$ $(45.9\%)$ & $34/50$ $(68.0\%)$ & $ 39/55$ $(70.9\%)$ & $51.6\%$  \\
        FTC        & $24/42$ $(57.1\%)$ & $21/37$ $(56.8\%)$ & $37/50$ $(74.0\%)$ & $ 37/55$ $(67.3\%)$ & $63.8\%$  \\
        FTC w/ WC  & $18/42$ $(42.9\%)$ & $19/37$ $(51.4\%)$ & $38/50$ $(76.0\%)$ & $ 40/55$ $(72.7\%)$ & $60.8\%$  \\
        \midrule
        w/o camera poses & $22/42$ $(52.4\%)$ & $20/37$ $(54.1\%)$ & $39/50$ $(78.0\%)$ & $ 37/55$ $(67.3\%)$ & $63.0\%$  \\
        w/o joint currents & $24/42$ $(57.1\%)$ & $22/37$ $(59.5\%)$ & $42/50$ $(84.0\%)$ & $ 41/55$ $(74.5\%)$ & $68.8\%$  \\
        \midrule
        \addlinespace[1.5pt]
        \rowcolor{green!30}
        \textbf{Ours} & $\mathbf{31/42}$ $(\mathbf{73.8}\textbf{\%})$ & $\mathbf{28/37}$ $(\mathbf{75.7}\textbf{\%})$ & $\mathbf{45/50}$ $(\mathbf{90.0}\textbf{\%})$ & $\mathbf{46/55}$ $(\mathbf{83.6} \textbf{\%})$ & $\mathbf{80.8}\textbf{\%}$  \\
        \bottomrule
    \end{tabular}
    \vspace{-3mm}
\end{table}

\begin{figure}[h!]
    \centering
    \begin{minipage}[t]{0.5\linewidth}
        \vspace{2pt}
        \centering
        \includegraphics[width=1.0\linewidth]{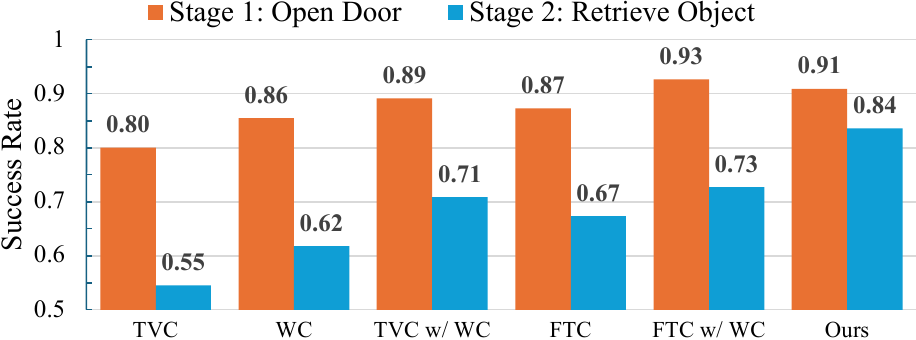}
        \captionof{figure}{\textbf{Stage-wise success rate of S4.} Stage 1 evaluates door opening, and Stage 2 evaluates object retrieval. } 
        \label{task4_results}
    \end{minipage}
    \hfill
    \begin{minipage}[t]{0.48\linewidth}
        \vspace{2pt}
        \centering
        \captionof{table}{Comparisons of system latency, computational cost, and average success rate between ours (\textit{FTC w/ TVC}) and the baseline (\textit{TVC w/ WC}).}
        \label{system_stat}
        \setlength{\tabcolsep}{3pt}
        \renewcommand{\arraystretch}{1.05}
        \scriptsize
        \begin{tabular}{lcc}
        \toprule
        \textbf{Metric} & \textbf{Ours} & \textbf{Baseline} \\
        \midrule
        Data Acquisition Time / Frame  & 35.2\,ms & 34.4\,ms \\
        Policy FLOPs & 215.01\,GFLOPs & 79.00\,GFLOPs \\
        End-to-End System Speed & 10.2\,FPS & 13.2\,FPS \\
        \midrule
        Average Success Rate & 80.8\% & 51.6\% \\
        \bottomrule
        \end{tabular}
    \end{minipage}
\end{figure}

\textbf{Evaluation Result}\quad As shown in Table~\ref{quantitative_results}, our method achieves success rates of $73.8\%$, $75.7\%$, $90.0\%$, and $83.6\%$, respectively, with an average success rate of $80.8\%$, outperforming all learned-policy baselines across these tasks. The results show that fingertip visual observations are particularly effective in occluded and spatially constrained manipulation scenarios. In S1 and S2, wrist-mounted and third-view cameras often fail to provide reliable local cues during insertion or pre-contact alignment, leading to collisions, misses, or object knock-offs. In S3 and S4, external views are frequently blocked by the curtain, cabinet structure, hand, or object itself, while fingertip views provide closer and more task-relevant visual feedback for grasping and retrieval. Compared with FTC-only baselines, our method benefits from combining fingertip observations with global scene context, which improves both approach-phase localization and contact-phase manipulation. For S4, Fig.~\ref{task4_results} further reports the stage-wise success rates for door opening and object retrieval. Although most methods with local perception perform quite well in Stage 1 with moderate occlusion, their performance degrades substantially in Stage 2, where severe in-cabinet occlusion requires close-range, multi-view fingertip perception for reliable object localization, grasping, and retrieval. These results demonstrate that hand-centric multi-view perception provides a more robust sensing strategy for long-horizon dexterous manipulation under severe occlusion. More detailed analysis and category-level results are provided in Appendix~\ref{app:analysis} and~\ref{category_results}, respectively.

\subsection{System Latency and Bandwidth} 
To assess trade-offs introduced by multi-camera inputs, we report the data acquisition time, end-to-end system speed, and policy computational cost, compared with the standard two-camera baseline (\textit{TVC w/ WC}), as shown in Table~\ref{system_stat}. Although our method processes more image streams, the additional overhead remains manageable and still supports real-time dexterous manipulation, while yielding a $56.6\%$ relative improvement in success rate.

\subsection{Ablation Studies} 
To illustrate the importance of camera pose encoding and finger joint current encoding modules of FingerViP, we conduct two ablation experiments across the four tasks, with the same training dataset and evaluation scenarios: (i) \textit{w/o camera poses}, where the policy is trained without camera pose inputs; (ii) \textit{w/o joint currents}, where the policy is trained without finger joint current inputs.

As shown in Table~\ref{quantitative_results}, our results reveal the critical importance of camera poses and finger current signals for learning a dexterous manipulation policy. Specifically, for contact-rich tasks such as button pressing and object grasping, removing finger current inputs significantly reduces success rates, highlighting the essential role of contact in understanding physical interactions between the robotic hand and objects. On the other hand, for tasks that require continuous and dexterous hand motion, such as unstable-support stick retrieval, incorporating camera pose encoding for dynamically moving fingertip cameras notably enhances multi-view perception, which improves the policy’s ability to execute accurate movements to center and conform the stick. These findings emphasize the necessity of both finger currents and camera poses to achieve robust dexterous manipulation. Additional analysis of finger joint-current signals is provided in Appendix~\ref{sec:joint_current_analysis}, where the current trajectories exhibit distinct variations across different interaction phases.

\begin{wraptable}{R}{0.50\linewidth}
\vspace{-2mm}
\centering
\setlength{\tabcolsep}{2pt}  
\scriptsize
\begin{tabular}{lccc}
\toprule
\textbf{Visual Encoder} & \textbf{1st Action} & \textbf{First $4$ Actions} & \textbf{First $8$ Actions} \\
\midrule
Frozen CLIP       & \colorbox{gray!30}{$\mathbf{0.0200}$} & \colorbox{gray!30}{$\mathbf{0.0272}$} & $0.0377$ \\
Frozen DINO-v2     & $0.0214$ & $0.0282$ & $0.0374$ \\
Fine-tuned CLIP   & $0.0232$ & $0.0303$ & $0.0396$ \\
Fine-tuned DINO-v2 & $0.0209$ & $0.0278$ & \colorbox{gray!30}{$\mathbf{0.0371}$}  \\
CLIP from Scratch & $0.0225$ & $0.0306$ & $0.0428$    \\
DINOv2 from Scratch & $0.0215$ & $0.0294$ & $0.0406$    \\
\bottomrule
\end{tabular}
\caption{Comparison of visual encoders in terms of MAE over the first 8 executed actions on 20 real-world evaluation trajectories. Lower is better.}
\label{encoder_compare}
\vspace{-2mm}
\end{wraptable}

To further examine the design choice of the visual encoder, we conduct an ablation study comparing frozen CLIP with fine-tuned CLIP, DINO-v2, and training from scratch. We evaluate these variants using action prediction accuracy, which is more aligned with our downstream task than isolated visual recognition metrics. As shown in Table~\ref{encoder_compare}, the frozen CLIP encoder achieves the best accuracy for the first action and the first four executed actions and delivers competitive results for the first eight actions without fine-tuning efforts, making it a practical and effective choice for our system.

We further evaluate fingertip-camera configurations in Appendix~\ref{sec:camera_ablation}, where the configuration with cameras on all five fingertips achieves the lowest action-prediction error across all four tasks compared with reduced-camera variants.

\section{Conclusions}

In this paper, our key insight is to bring clear visual perception directly to the fingertips and present FingerViP, a learning system that combines fingertip visual hardware with learning-based control for real-world dexterous manipulation. FingerViP significantly outperforms learned-policy baselines on four challenging real-world dexterous manipulation tasks while maintaining practical real-time performance, highlighting the importance of multi-view fingertip visual perception. Overall, our system offers a scalable multi-view visual perception approach for multi-fingered dexterous manipulation.

\begin{wrapfigure}{R}{0.52\linewidth}
    \centering
    \vspace{-3mm}
    \includegraphics[width=1.0\linewidth]{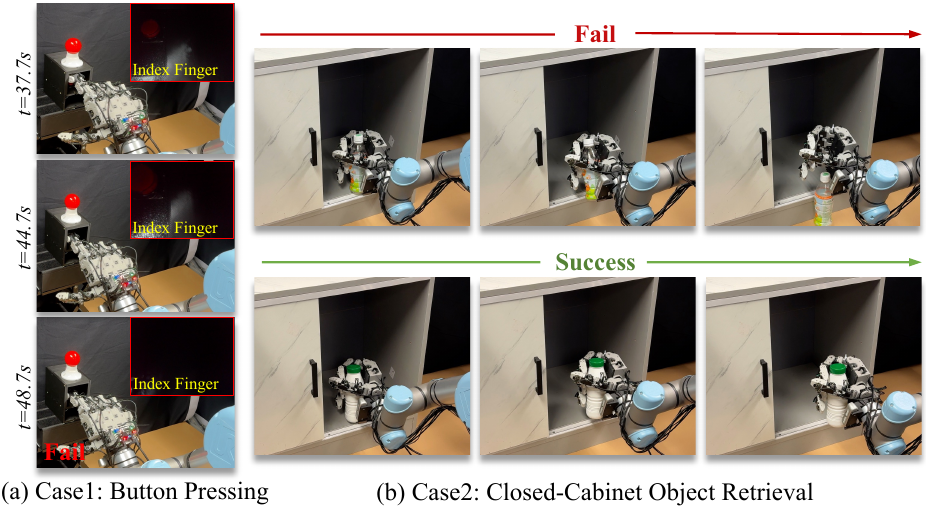}
    \caption{\textbf{Representative failure cases.} (a) Button pressing under no illumination. (b) Closed-cabinet retrieval of a slender, slippery object.}
    \label{failure}
    \vspace{-3mm}
\end{wrapfigure}

\section{Limitations and Future Work}

\textbf{Limitations}\quad We analyze two representative failure cases of FingerViP (Fig.~\ref{failure}). \textit{Case 1:} In S1, when pressing a non-illuminated button, the fingertip view becomes nearly dark after insertion, making the button hard to perceive and localize, leading to repeated misalignment and task failure. \textit{Case 2:} In S4, we observed that FingerViP often fails on a slender, slippery juice bottle that tends to slip out of the hand during grasping and retrieval. Nevertheless, FingerViP performs well on a similar-sized white bottle with a rough, matte surface. This suggests that the current hand surface may provide insufficient friction, making it easier for slender and slippery objects to slip out under grasping forces. Our experiments focus on multi-finger manipulation under confined and severely occluded conditions, and do not cover broader dexterous capabilities such as complex in-hand reorientation. Extending fingertip visual perception to these scenarios remains future work.

\textbf{Future work}\quad We will pursue a more compact and robust fingertip module design, including further miniaturization, integrated illumination for low-light scenarios, thermal management for long-term camera operation, and enhanced contact surfaces for stable grasping. We will also systematically investigate fingertip-camera placement and image-resolution trade-offs, together with compact tactile sensing for complementary fine-grained contact information. We will also explore asynchronous execution strategies~\cite{wu2024tidybot++,black2026real} to mitigate the effects of policy-inference latency on control responsiveness. Furthermore, we aim to exploit the calibrated multi-view fingertip setup for local 3D reconstruction with known camera poses, enabling geometry-aware state estimation and policy learning.
\clearpage
\acknowledgments{This work was supported in part by the Multi-Scale Medical Robotics Centre, AIR@InnoHK. We would like to thank CRML@CUHK for generously providing the UR5e robotic arm, and Dezhao Guo for his assistance. We are also grateful to Professor Shing Shin Cheng for his support with a robotic arm for testing. We also thank Yanlin Chen, Jianghua Chen, Pui Hin NG, Ziliang Feng, and Yongjun Yan for their helpful feedback and fruitful discussions.}


\bibliography{references}  

\clearpage

\section*{Appendix}
\appendix
\section{Details of Experiments}
\subsection{Test Scenarios of Four Tasks}
\label{app:scenarios}
\begin{figure}[!h]
    \centering
    \includegraphics[width=1.0\linewidth]{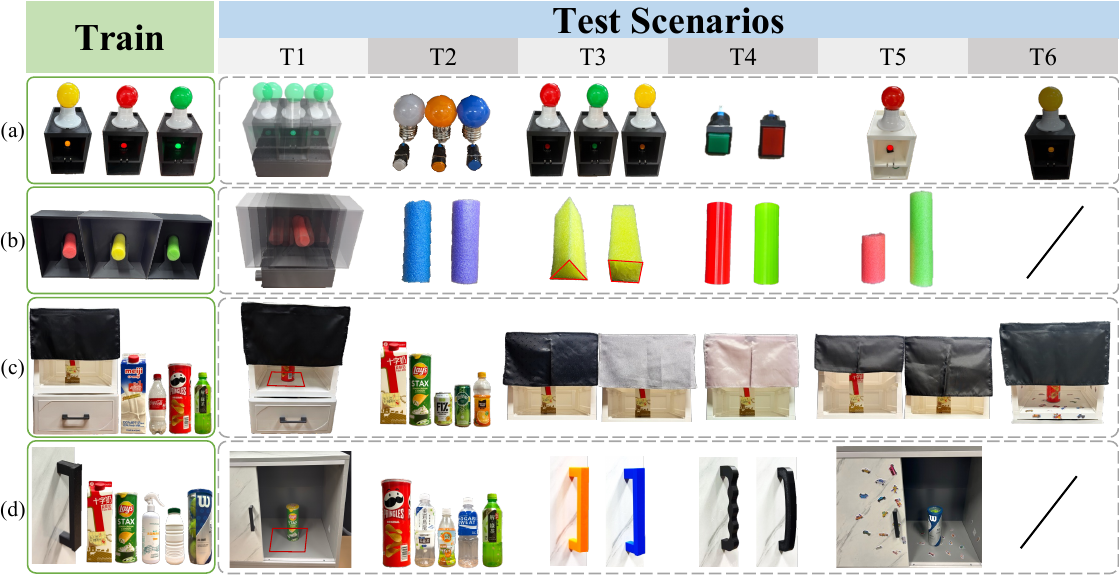}
    \caption{\textbf{Scenarios for Four Challenging Real-World Tasks.} Each row lists the various cases for both training and testing.}
    \label{test_scenarios}
\end{figure}

\textbf{Confined-Box Button Pressing (S1)}\quad We conduct \textbf{42} rollouts for each policy. As shown in Fig.~\ref{test_scenarios} (a), the test scenarios can be grouped into the following six categories: \textit{T1)} Variations in the initial button-box position (18 rollouts); \textit{T2)} Buttons in unseen colors (6 rollouts); \textit{T3)} Non-illuminated buttons (6 rollouts); \textit{T4)} Buttons in unseen shapes (6 rollouts); \textit{T5)} A white button box (3 rollouts); and \textit{T6)} Changes in ambient light intensity (3 rollouts).

\textbf{Unstable-Support Stick Retrieval (S2)}\quad We conduct \textbf{37} rollouts for each policy. As shown in Fig.~\ref{test_scenarios} (b), the test configurations can be grouped into five categories: \textit{T1)} Variations in the initial box positions (18 rollouts); \textit{T2)} Foam sticks in unseen colors (6 rollouts); \textit{T3)} Foam sticks in unseen shapes (6 rollouts); \textit{T4)} Varied stick material (thermoplastic polyurethane, TPU; 3 rollouts); and \textit{T5)} Foam sticks with unseen lengths (4 rollouts).

\textbf{Curtain-Occluded Object Retrieval (S3)}\quad We conduct \textbf{50} rollouts for each policy. As shown in Fig.~\ref{test_scenarios} (c), the test configurations can be grouped into six categories: T1) Variations in initial object poses (20 rollouts); T2) Unseen target objects (10 rollouts); T3) Curtains with unseen materials (5 rollouts); T4) Curtain in unseen color (5 rollouts); T5) Curtains with unseen lengths (5 rollouts); and \textit{T6)} Unseen cabinet wall appearance (stickers; 5 rollouts).

\textbf{Closed-Cabinet Object Retrieval (S4)}\quad We conduct \textbf{55} rollouts for each policy. As shown in Fig.~\ref{test_scenarios} (d), the test configurations can be grouped into five categories: \textit{T1)} Variations in object initial poses, seen during training (25 rollouts); \textit{T2)} Unseen target objects (15 rollouts); \textit{T3)} Cabinet handle with unseen shapes (5 rollouts); \textit{T4)} Cabinet handle in unseen colors (5 rollouts); and \textit{T5)} Unseen cabinet wall appearance (stickers; 5 rollouts).

\subsection{Per-Category Quantitative Results}
\label{category_results}
We further report the category-level success rate (SR) of FingerViP across the testing scenarios for the four tasks in Table~\ref{trial_success}.
\begin{table}[!h]
    \centering
    \caption{Category-level SR across all testing scenarios for four tasks.}
    \label{trial_success}
    \footnotesize
    \begin{tabular}{lccccccc}
        \toprule
        \textbf{Task} & \textbf{T1} & \textbf{T2} & \textbf{T3} & \textbf{T4} & \textbf{T5} & \textbf{T6} & \textbf{Total}\\
        \midrule
        S1 & 17/18 & 4/6 & 4/6 & 3/6 & 1/3 & 2/3 & 31/42 \\
        S2 & 14/18 & 4/6 & 5/6 & 2/3 & 3/4 & -- & 28/37 \\
        S3 & 18/20 & 7/10 & 5/5 & 5/5 & 5/5 & 5/5 & 45/50 \\
        S4 & 22/25 & 13/15 & 4/5 & 3/5 & 4/5 & -- & 46/55 \\
        \bottomrule
    \end{tabular}
\end{table}

\subsection{Fingertip Camera Configuration Ablation}
\label{sec:camera_ablation}
Our system equips all five fingertips with cameras to provide a task-agnostic whole-hand perception interface. Since the informativeness of individual views varies across tasks and interaction phases, we retain all five views rather than selecting a task-specific subset. To assess this design choice, we compare the full configuration with cameras on all five fingertips against reduced-camera variants.

\begin{table}[!h]
    \centering
    \footnotesize
    \caption{Fingertip camera configuration ablation: Action-prediction MAE ($\times10^{-3}$) over the first 8 executed actions across $80$ evaluation trajectories. Lower is better.}
    \label{camera_ablation}
    \begin{tabular}{lccccc}
        \toprule
        \textbf{Camera Configuration} & \textbf{S1} & \textbf{S2} & \textbf{S3} & \textbf{S4}  & \textbf{Avg.} \\
        \midrule
        TVC w/ Thumb + Index          & $35.54$ & $34.55$ & $35.09$ & $30.30$ & $33.87$  \\
        TVC w/ Thumb + Index + Little & $32.00$ & $32.10$ & $33.67$ & $28.53$ & $31.58$  \\
        TVC w/ Index + Middle + Ring  & $30.79$ & $32.30$ & $35.63$ & $29.38$ & $32.03$  \\
        TVC w/ Five Fingertips (Ours)    & $\mathbf{29.55}$ & $\mathbf{30.81}$ & $\mathbf{33.21}$ & $\mathbf{27.90}$ & $\mathbf{30.37}$  \\
        \bottomrule
    \end{tabular}
\end{table}

As shown in Table~\ref{camera_ablation}, the configuration using cameras on all five fingertips achieves the lowest action-prediction error across all four tasks, with an average MAE of $30.37\times10^{-3}$. This suggests that different fingertip views provide complementary information and that a fixed reduced subset may not generalize consistently across manipulation scenarios. The full configuration therefore provides a robust, task-agnostic whole-hand perception interface without task-specific view selection.

\subsection{Detailed Analysis of Experimental Results for Each Task}
\label{app:analysis}
\textbf{Results Analysis of S1}\quad The \textit{WC} policy performs the worst (Table~\ref{quantitative_results}), due to the heavily occluded wrist view when the other four fingers are curled, making it difficult to complete the task. The \textit{TVC} policy often fails to localize the box opening and internal button precisely under occlusion, leading to collisions with the box during insertion. The \textit{TVC w/ WC} baseline, despite leveraging two views, still suffers from the occlusion problem. Compared to ours, the \textit{FTC} baseline performs worse during the approach phase with large arm-hand motions, which is consistent with the limited global scene awareness without a true global view (i.e., a third-view camera). The \textit{FTC w/ WC} baseline degrades further compared to \textit{FTC}. In this task, the insertion-relevant local cues come primarily from the index-finger view, as the other fingers are fully curled and their fingertip views are largely self-occluded. The wrist view is also blocked by the curled fingers, providing limited useful information.

\textbf{Results Analysis of S2}\quad The \textit{TVC} baseline can roughly localize the stick, but struggles to keep it centered during the approach and pre-contact phases, resulting in frequent knocking-offs due to the absence of reliable local visual feedback. The \textit{WC} policy often fails to localize the stick precisely and to maintain stable contact, leading to frequent misses or knock-offs. Although \textit{TVC w/ WC} leverages multi-view inputs, it still suffers from frequent knock-offs due to insufficient local perception. The \textit{FTC} policy is weaker during the approach phase because it lacks a true global perception (e.g., a static third-view). However, once it reaches the box, it tracks the stick center accurately, establishes stable contact, and completes the task. Adding a wrist view (\textit{FTC w/ WC}) does not resolve the localization issue and yields performance comparable to \textit{FTC}.

\textbf{Results Analysis of S3}\quad In this task, occlusion from the hanging curtain limits the effectiveness of the \textit{TVC} approach. When the object is placed deep inside or close to the cabinet’s left side and becomes fully hidden by the curtain and the cabinet wall, the hand is unable to reach the correct position, causing grasp failures. A single wrist-mounted camera (\textit{WC}) is also insufficient because the thumb can occlude the view during grasping, leading to unstable grasps. And without reliable global scene awareness, collisions with the cabinet can trigger the arm’s e-stop. Using both third-view and wrist-mounted cameras (\textit{TVC w/ WC}) reduces e-stop events, but limited local perception remains a bottleneck. The \textit{FTC} and \textit{FTC w/ WC} policies localize the object reliably, yet they still often collide with the cabinet bottom during retrieval, triggering e-stops and terminating execution.

\textbf{Results Analysis of S4}\quad During door opening, the \textit{TVC} policy may extend the fingers behind the door near the handle and get jammed, leading to opening failures. The object is often only partially visible and can become fully occluded with pose variations, resulting in a low retrieval success rate. The \textit{WC} policy supports door opening, but finger–handle misalignment can still occur. During in-cabinet grasping, the thumb may occlude the wrist camera, and without global context the door is sometimes not opened wide enough. During retrieval, the object may collide with the cabinet bottom or the door, causing failures or e-stops. Combining third-view and wrist views (\textit{TVC w/ WC}) helps open the door to the desired position, but grasping the object inside the cabinet remains unstable due to thumb occlusion of the wrist camera and object-induced occlusion of the wrist view. Fingertip-based policies (\textit{FTC} and \textit{FTC w/ WC}) perform well in door opening, but inadequate opening and collision-induced e-stops can still occur. This is consistent with limited global awareness of door opening state and bottom clearance during retrieval.

\subsection{Task Performance with Human Teleoperation}
We additionally report the performance of Human Teleoperation (HT) method, where the robot is directly controlled by a human operator using Apple Vision Pro, as shown in Table~\ref{ht_results}. We consider HT as an approximate upper-bound reference for manipulation performance rather than a learned-policy baseline, since it benefits from human perception and decision-making capabilities. Nevertheless, HT does not always achieve perfect performance in practice, and some trials fail due to unstable hand tracking, communication latency, and network instability during teleoperation.
\begin{table}[!h]
    \centering
    \vspace{-3mm}
    \caption{Performance of human teleoperation on four experimental tasks.}
    \label{ht_results}
    \small
    \setlength{\tabcolsep}{4pt} 
    \begin{tabular}{lccccc}
        \toprule
        \multirow{2}{*}{\textbf{Methods}}
        & \multirow{2}{*}{\makecell{Confined-Box\\Button Pressing}}
        & \multirow{2}{*}{\makecell{Unstable-Support\\Stick Retrieval}}
        & \multirow{2}{*}{\makecell{Curtain-Occluded\\Object Retrieval}}
        & \multirow{2}{*}{\makecell{Closed-Cabinet \\ Object Retrieval}}
        & \multirow{2}{*}{\makecell{\textbf{Average}\\ \textbf{Success Rate}}} \\
        &  &  &  &  &  \\
        \toprule 
        HT       & $42/42$ $(100.0\%)$ & $34/37$ $(91.9\%)$ & $48/50$ $(96.0\%)$ & $ 54/55$ $(98.2\%)$ & $96.5\%$  \\    
        \bottomrule
    \end{tabular}
    \vspace{-3mm}
\end{table}

\subsection{Finger Joint-Current Signal Analysis}
\label{sec:joint_current_analysis}
\begin{figure}[!h]
    \centering
    \includegraphics[width=0.55\linewidth]{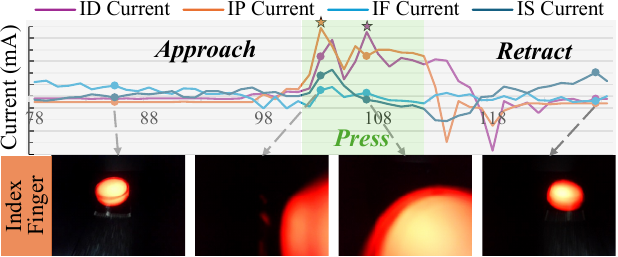}
    \caption{Index-finger joint current trajectories in S1, showing distinct current variations during the approach, press, and retract phases.}
    \label{fig:current_curve}
\end{figure}

To further characterize the per-finger joint-current signals used by our policy, we visualize the index-finger joint-current trajectories in S1. As shown in Fig.~\ref{fig:current_curve}, the current signals exhibit distinct patterns across the approach, press, and retract phases, reflecting phase-dependent changes in motor loading during interaction. These signals are encoded as contact-related cues rather than treated as calibrated force measurements.

\section{Teleoperation and Data Collection}
\label{app:tdc}
\vspace{-2mm}
\textbf{Teleoperation}\quad To efficiently collect human demonstrations, we employ a high-DoF teleoperation interface~\cite{park2024avp} that captures human hand poses with an Apple Vision Pro and enables real-time control of the robotic arm and hand. Specifically, an Apple Vision Pro tracks the operator’s hand at a frequency of \SI{60}{\hertz} and streams $21$ hand keypoints and a $7$-D wrist pose (i.e., 3D position and quaternion orientation) over UDP protocol. To align each human keypoint with its corresponding robot hand link, a per-phalangeal segment geometric calibration is applied to reduce local kinematic discrepancies and mitigate global-scale distortions. The human hand motions are retargeted by solving an optimization problem with an objective consisting of a conformal-aligned constraint, a contact-aware coupling constraint, and temporal smoothness~\cite{wan2025rapid}. The joint angles of a UR5e arm are computed through inverse kinematics based on the wrist pose, enabling coordinated arm–hand control. The control system runs at \SI{100}{\hertz} and publishes $26$ joint commands to the arm and hand.

\textbf{Data Collection}\quad In this work, we collect (i) target joint commands, (ii) robot proprioception: arm-hand joint positions for all 26 DoFs, and hand joint currents for the 20 hand DoFs, and (iii) multi-view RGB observations from six cameras: five fingertip cameras and one third-view RGB-D camera (RealSense D435i). Additional synchronized wrist-view observations are recorded for baseline comparisons. These multi-modal signals run at different frequencies. Specifically, all cameras operate at \SI{30}{\hertz} with a resolution of $640 \times 480$, and all target joint commands are received at \SI{100}{\hertz}. However, precise temporal synchronization of multi-modal data is critical for collecting data and training policies from demonstration data for dexterous manipulation tasks. 

To address this, each fingertip camera is captured in an independent thread and synchronized via a global threading event, ensuring that all cameras are initialized before acquisition and that frames are temporally aligned across views. Each fingertip camera maintains a frame-queue buffer for parallel capture with low-latency access to the latest observation (\qty{6.7}{\milli\second} average latency across five fingertip cameras). In addition, dedicated threads are used to collect robot proprioception and synchronized RGB streams. Once control commands are dispatched by the main controller, proprioceptive data and all RGB streams are recorded simultaneously under the same strategy. Finally, the resulting data are compressed and saved in Zarr format for post-processing.

\vspace{-2mm}
\subsection{Dataset Statistics}

We collected $1012$ real-world demonstration episodes through teleoperation for the four training tasks shown in Fig.~\ref{test_scenarios}. The training objects are shown in the left column, while the corresponding test scenarios T1--T6 are shown on the right to evaluate generalization to changes in object appearance, color, layout, distractors, and occlusion conditions. The detailed dataset statistics are as follows:
\begin{itemize}[leftmargin=1.0em, itemsep=1.0pt, topsep=2pt]
    \item \textbf{S1:} The training dataset of S1 contains three button boxes with yellow, red, and green buttons, as shown in Fig.~\ref{test_scenarios} (a). We collected $255$ demonstration episodes in total, with an average duration of \SI{7.2}{\second} per episode.
    
    \item \textbf{S2:} The training dataset of S2 contains three foam craft sticks in red, yellow, and green, as shown in Fig.~\ref{test_scenarios} (b). We collected $256$ demonstration episodes in total, with an average duration of \SI{6.1}{\second} per episode.
    
    \item \textbf{S3:} The training dataset of S3 contains four target objects placed inside a cabinet behind the hanging curtain, as shown in Fig.~\ref{test_scenarios} (c). We collected $245$ demonstration episodes in total, with an average duration of \SI{8.5}{\second} per episode.
    
    \item \textbf{S4:} The training dataset of S4 contains five target objects placed inside a cabinet, as shown in Fig.~\ref{test_scenarios} (d). We collected $256$ demonstration episodes in total, with an average duration of \SI{13.5}{\second} per episode.
\end{itemize}

\section{Additional Implementation Details}
\label{app:implementation}

\subsection{Hardware Setup Details}
An Apple Vision Pro was used to collect teleoperation commands. System control, data collection, and policy evaluation were performed on a desktop PC equipped with an Intel i7-12700K CPU, an NVIDIA RTX 4080 GPU with 16 GB memory, and 32 GB RAM. To support reliable multi-camera streaming, the PC was further equipped with a PCIe USB expansion card, which provided an additional 20 Gbps of dedicated bandwidth for seven camera streams. Across 1,012 teleoperation demonstrations and 1,472 evaluation rollouts, no fingertip camera module or FPC replacement was required, indicating that the current hardware design was sufficiently robust for our experiments.

\subsection{Robot Initial State}
\label{app:init_pose}

\textbf{Robot arm}\quad Before each episode, the robot arm (i.e., UR5e) is reset to a predefined initial joint configuration to ensure consistent data collection and policy execution. Since different tasks are conducted under different workspace layouts, we use two task-specific initial pose configurations summarized in Table~\ref{tab:init_pose}.

\begin{table}[!h]
    \centering
    \caption{Initial robot arm joint configurations used in the experiments.}
    \label{tab:init_pose}
    \small 
    \setlength{\tabcolsep}{4pt} 
    \begin{tabular}{lcccccc}
        \toprule
        \textbf{Task} & \textbf{Shoulder pan} & \textbf{Shoulder lift} & \textbf{Elbow} & \textbf{Wrist 1} & \textbf{Wrist 2} & \textbf{Wrist 3} \\
        \midrule
        S1 / S2 
        & 0.0552 & -1.7843 & 2.4377 & -1.2578 & 1.5504 & -0.0036 \\
        S3 / S4 
        & -0.2230 & -1.2892 & 2.7728 & -2.2508 & 1.3989 & 0.1320 \\
        \bottomrule
    \end{tabular}
    \vspace{-2mm}
\end{table}

\textbf{Robot hand}\quad For the 20-DoF dexterous hand, all joint angles are initialized to zero for all tasks.

\subsection{Policy Training and Testing Details}
The policy is trained on an NVIDIA RTX 4090 GPU, taking approximately 37 hours on average. An overview of important hyperparameters used for training and testing is provided in Table~\ref{hyperparameter}.
\begin{table}[!h]
    \centering
    \begin{tabular}{c|c}
        \toprule
        Hyperparameter & Value \\
        \midrule
        Observation horizon & 2 \\
        Action prediction horizon & 16 \\
        Observation resolution & $224\times224$   \\
        Number of attention heads & 8 \\
        Finger current projection dim & 128 \\
        Camera orientation projection dim & 128\\
        Camera position projection dim & 128 \\
        Joint angle projection dim & 768 \\
        Optimizer & AdamW \\
        Optimizer momentum & $\beta_1, \beta_2 = 0.95, 0.999$ \\
        Learning rate & 3e-4 \\
        Batch size & 64 \\
        Weight decay & 1e-6 \\
        Denoising steps & 16 \\
        Action execution horizon & 8 \\
        \bottomrule
    \end{tabular}
    \caption{Hyperparameters for policy training and testing.}
    \label{hyperparameter}
    \vspace{-2mm}
\end{table}

\subsection{Fingertip Camera Motion Artifacts}
Due to the dynamic motion of fingertip cameras, visual artifacts such as image blur are unavoidable during manipulation. To reduce the impact of motion effects on policy training, we manually filter demonstrations and remove a few severely blurred episodes to ensure higher-quality training data. In addition, to improve robustness to such visual degradation, we apply standard visual augmentations, including \textit{ColorJitter} and \textit{GaussianBlur}, during training to improve robustness to moderate degradation at testing. These augmentations randomly perturb image brightness, contrast, saturation, and hue, and randomly apply Gaussian blurring with sampled parameters. Note that the \texttt{GaussianBlur} is only implemented for the fingertip cameras.

The configurations of \textit{ColorJitter} transformation and \textit{GaussianBlur} transformation used for policy training are shown in Table~\ref{colorjitter} and Table~\ref{gaussianblur}.

\begin{table}[!h]
    \centering
    \begin{minipage}[t]{0.48\linewidth}
        \vspace{0pt}
        \centering
        \begin{tabular}{cc}
            \toprule
            Configuration & Value \\
            \midrule
            Brightness factor& 0.3 \\
            Contrast factor& 0.4 \\
            Saturation factor& 0.5 \\
            Hue factor& 0.08 \\
            \bottomrule
        \end{tabular}
        \caption{Configurations of \textit{ColorJitter} transformation.}
        \label{colorjitter}
    \end{minipage}
    \begin{minipage}[t]{0.48\linewidth}
        \vspace{0pt}
        \centering
        \begin{tabular}{cc}
            \toprule
            Configuration & Value \\
            \midrule
            Gaussian kernel size & 3 \\
            Standard deviation & [0.1, 0.3] \\
            \bottomrule
        \end{tabular}
        \caption{Configurations of \textit{GaussianBlur} transformation.}
        \label{gaussianblur}
    \end{minipage}
    \vspace{-6mm}
\end{table}

\section{Visualization of Multi-View Camera Observations}
\begin{figure}[!h]
    \centering
    \includegraphics[width=1.0\linewidth]{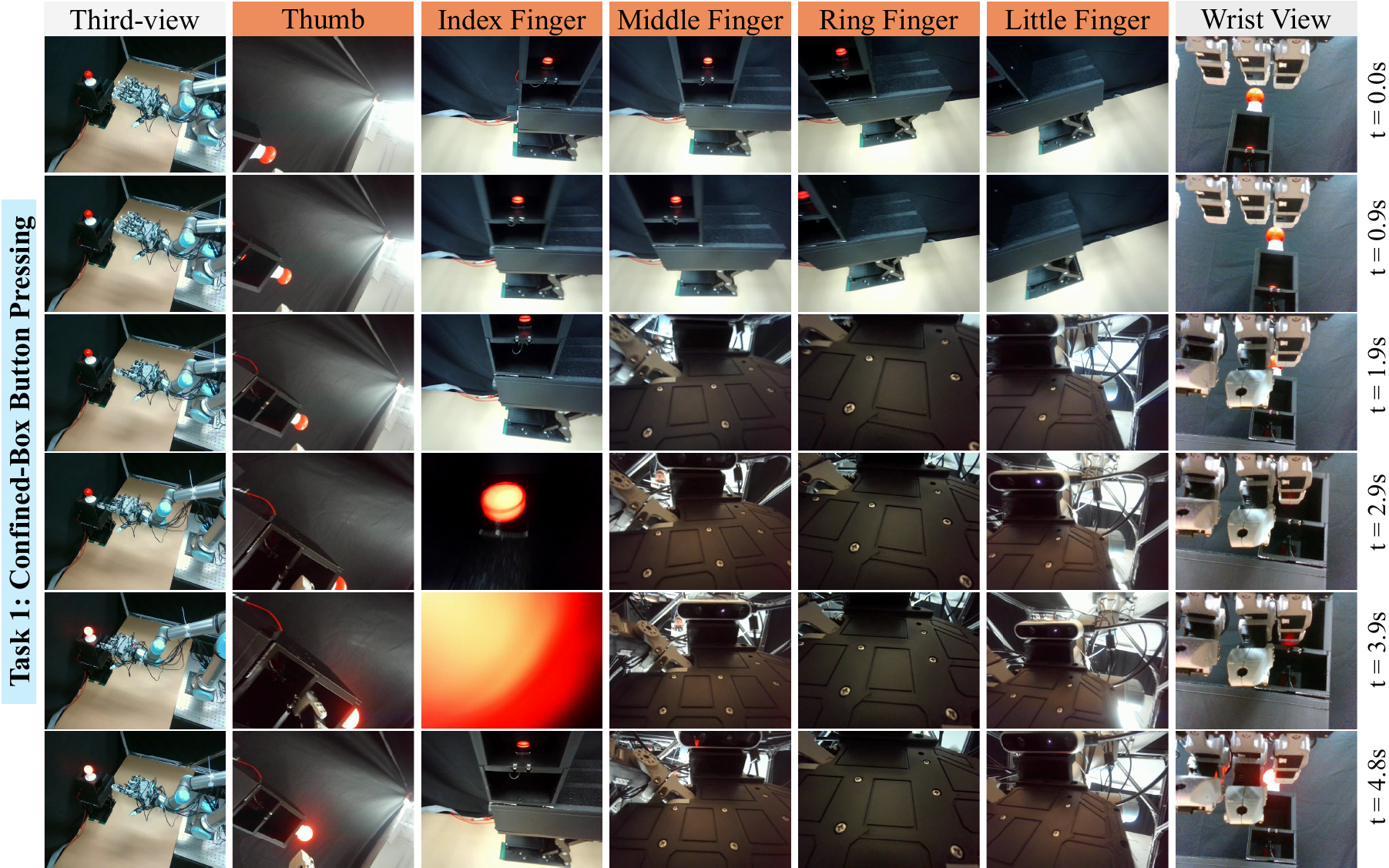}
    \caption{Visualizations of synchronized multi-view RGB observations in S1: Confined-Box Button Pressing.}
    \label{task1_allviews}
\end{figure}

\begin{figure}[!h]
    \centering
    \includegraphics[width=1.0\linewidth]{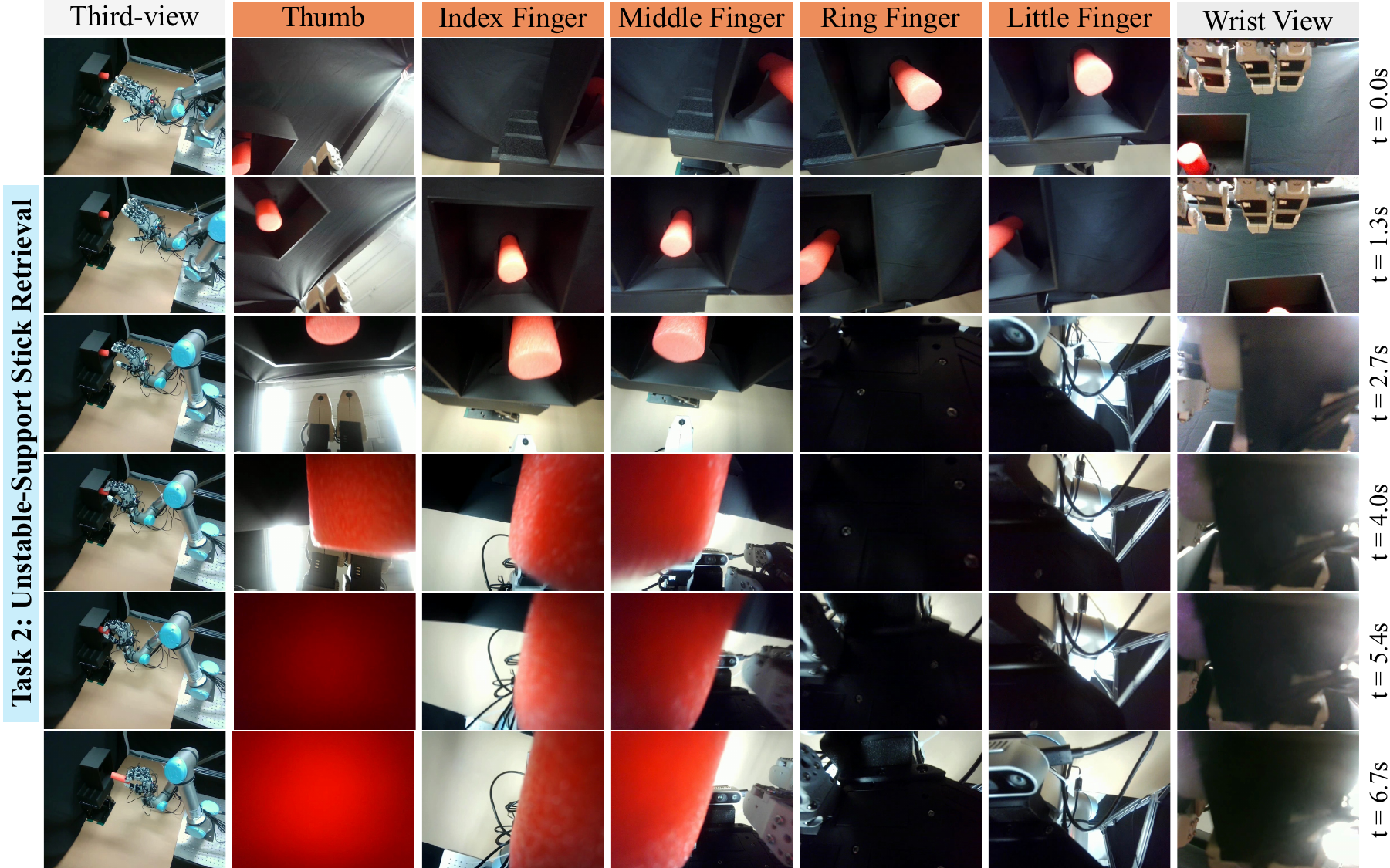}
    \caption{Visualizations of synchronized multi-view RGB observations in S2: Unstable-Support Stick Retrieval.}
    \label{task2_allviews}
\end{figure}

\begin{figure}[!h]
    \centering
    \includegraphics[width=1.0\linewidth]{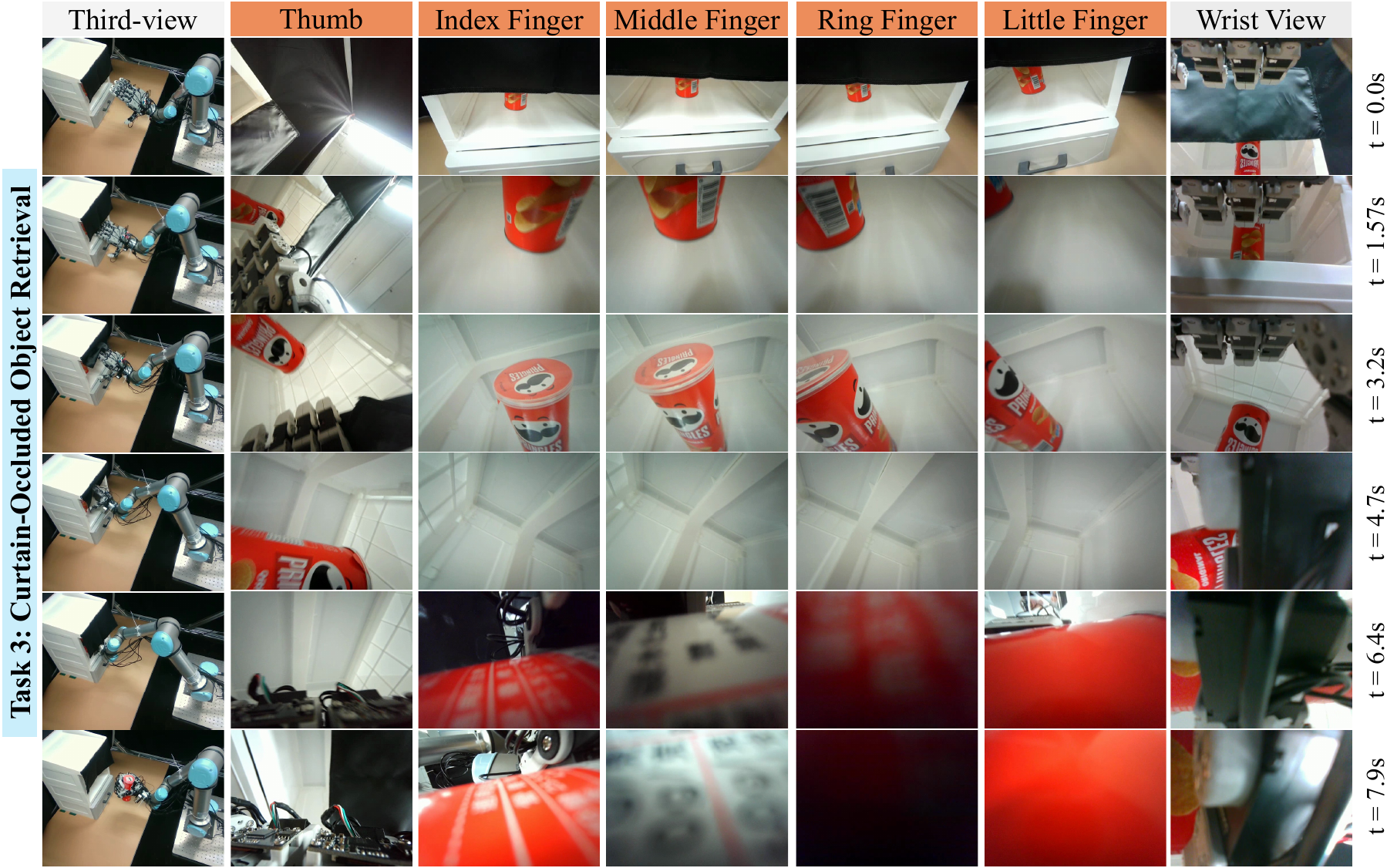}
    \caption{Visualizations of synchronized multi-view RGB observations in S3: Curtain-Occluded Object Retrieval.}
    \label{task3_allviews}
\end{figure}

\begin{figure}[!h]
    \centering
    \includegraphics[width=1.0\linewidth]{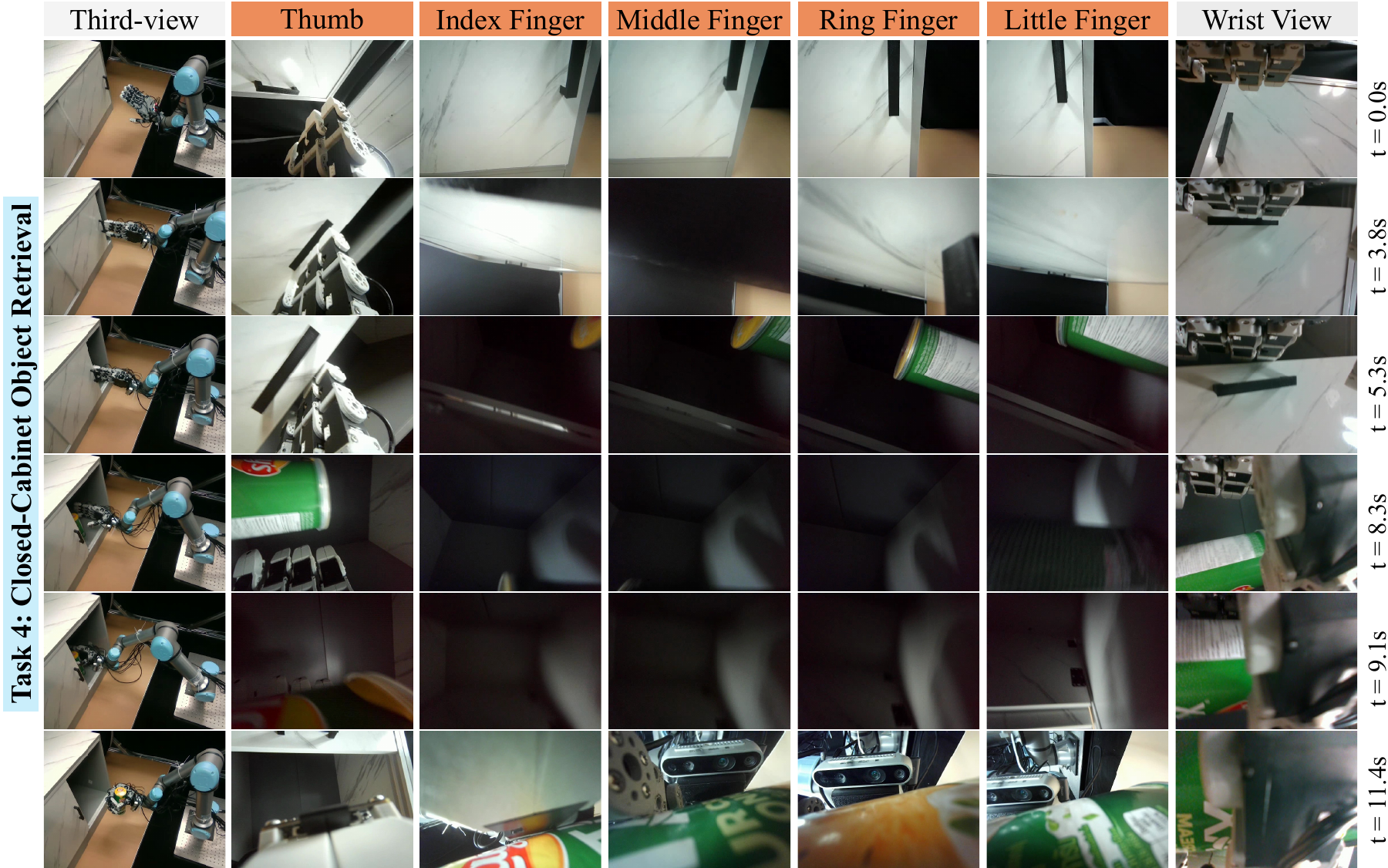}
    \caption{Visualizations of synchronized multi-view RGB observations in S4: Closed-Cabinet Object Retrieval.}
    \label{task4_allviews}
\end{figure}

Here, we visualize the RGB observations from the collected trajectories for each experimental task, including five fingertip cameras and one third-view camera. For completeness, the synchronized wrist-view observations recorded for baseline comparisons are also shown. As shown in Fig.~\ref{task1_allviews}, Fig.~\ref{task2_allviews}, Fig.~\ref{task3_allviews}, and Fig.~\ref{task4_allviews}, the fingertip cameras provide semantically rich, hand-centric visual feedback from multiple contact-relevant viewpoints during the approach and pre-contact phases, enabling the policy to capture object geometry, local interactions, and occluded regions that are difficult to observe from an external camera alone. 
 
The visualizations also support our design choice of using five fingertip cameras. As shown in the visualized trajectories, the task-relevant fingers can vary across different tasks and manipulation phases. Therefore, the five fingertip cameras are not intended to imply that all fingers are strictly or equally necessary for every task or every phase. Instead, the fingertip-camera module is designed to serve as a task-agnostic visual perception interface for a multi-finger dexterous hand. Since FingerViP is formulated as a whole-hand learning system, it aggregates visual features, camera poses, and joint-current information from all fingers, as described in Sec.~\ref{sec:policy}. This design enables FingerViP to naturally adapt to varying active-finger subsets without being tied to a fixed finger configuration.

\end{document}